%% file: main.tex
\documentclass{article}
\usepackage{ijcai25}

\usepackage{times}
\usepackage{soul}
\usepackage{url}
\usepackage[hidelinks]{hyperref}
\usepackage[utf8]{inputenc}
\usepackage[small]{caption} 
\usepackage{graphicx}
\usepackage{amsmath}
\usepackage{amsthm}
\usepackage{amssymb}
\usepackage{booktabs}
\usepackage{algorithm}
\usepackage{algorithmic}
\usepackage{multicol}
\usepackage{multirow}
\usepackage[switch]{lineno}
\usepackage{amsfonts}

\usepackage{subcaption}
\usepackage{makecell}
\usepackage{tcolorbox}
\usepackage{xcolor}

\newcommand\blfootnote[1]{%
	\begingroup
	\renewcommand\thefootnote{}\footnote{#1}%
	\addtocounter{footnote}{-1}%
	\endgroup
}

\usepackage{todonotes}

\usepackage{pifont}
\newcommand{\cmark}{\ding{51}}%
\newcommand{\xmark}{\ding{55}}%

\usepackage{tikz}

\title{Vertical Federated Learning in Practice: The Good, the Bad, and the Ugly}

\author{
Zhaomin Wu$^1$\and
Zhen Qin$^3$\and
Junyi Hou$^1$\and
Haodong Zhao$^4$\and
Qinbin Li$^5$\and\\
Bingsheng He$^1$\and
Lixin Fan$^2$
\affiliations
$^1$National University of Singapore, Singapore,
$^2$WeBank, China,
$^3$Zhejiang University, China\\
$^4$Shanghai Jiao Tong University, China,
$^5$Huazhong University of Science and Technology, China
\emails
zhaomin@nus.edu.sg,
zhaohaodong@sjtu.edu.cn,
zhenqin@zju.edu.cn,
junyi.h@comp.nus.edu.sg,
qinbin@hust.edu.cn,
hebs@comp.nus.edu.sg,
lixinfan@webank.com
}

\begin{document}

\maketitle

\begin{abstract}
Vertical Federated Learning (VFL) is a privacy-preserving collaborative learning paradigm that enables multiple parties with distinct feature sets to jointly train machine learning models without sharing their raw data. Despite its potential to facilitate cross-organizational collaborations, the deployment of VFL systems in real-world applications remains limited. To investigate the gap between existing VFL research and practical deployment, this survey analyzes the real-world data distributions in potential VFL applications and identifies four key findings that highlight this gap. We propose a novel data-oriented taxonomy of VFL algorithms based on real VFL data distributions. Our comprehensive review of existing VFL algorithms reveals that some common practical VFL scenarios have few or no viable solutions. Based on these observations, we outline key research directions aimed at bridging the gap between current VFL research and real-world applications. 
\end{abstract}

\section{Introduction}\label{sec:introduction}

\input{sections/introduction}

\section{Overview of Vertical Federated Learning}\label{sec:overview}
\input{sections/overview}

\section{VFL Data Distributions}\label{sec:scenario}
\input{sections/scenario}

\section{Taxonomy of Algorithms}\label{sec:system}
\input{sections/system}

\section{Discussion}\label{sec:future}
\input{sections/future}

\section{Conclusion}\label{sec:conclusion}
\input{sections/conclusion}

\bibliographystyle{named}
\bibliography{references-short}

\end{document}

%% file: sections/introduction.tex
\blfootnote{Preprint. Under Review.}

With the exhaustion of high-quality publicly available data \cite{villalobos2022will} and the increasing stringency of privacy regulations such as the European Union’s General Data Protection Regulation (GDPR)\footnote{https://gdpr-info.eu/}, federated learning \cite{mcmahan2017communication} has emerged as a promising solution for training machine learning models across multiple parties without sharing their sensitive data. Federated learning is typically classified into horizontal federated learning (HFL) and vertical federated learning (VFL) based on the data distribution among the parties \cite{yang2019federated}. In HFL, parties share the same feature set but have different samples, whereas in VFL, parties possess different feature sets but the same sample set.

In recent years, VFL has gained significant interest in industrial collaborations, as companies holding different features often complement and benefit from each other \cite{li2020review}. These companies typically operate in different domains, such as e-commerce and banking, which reduces conflicts of interest and fosters a greater willingness to collaborate. Especially in the era of large language models, collaboration between private domains with different features can create comprehensive foundation models \cite{zheng2023input}. Driven by high demand in industrial applications, many VFL systems have emerged \cite{liu2021fate,li2023fedtree}. However, despite this growing interest, very few VFL frameworks or systems have been deployed in real-world applications, as noted by recent studies \cite{khan2022vertical,ye2024vertical}. This gap between research and application is crucial for the direction of future VFL research.

Existing VFL surveys primarily emphasize algorithmic perspectives while overlooking real-world data distributions in practical applications. Surveys such as \cite{liu2024vertical,khan2022vertical,cui2024survey,ye2024vertical} present taxonomies of VFL algorithms, categorizing contributions by performance, efficiency, communication, and privacy. Additionally, surveys including \cite{liu2024label,li2023vertical,yu2024survey} focus on privacy issues, exploring attack and defense mechanisms. Although recent research \cite{nock2021impact,wu2022coupled} highlights that real-world VFL data is less ideal than current experimental settings, there remains a lack of systematic investigation into real-world VFL data distributions. This gap hinders the effective bridging between theoretical research and practical applications.

In this survey, we explore the data distribution of potential VFL applications using a recent real-world database corpus, WikiDBs \cite{vogel2024wikidbs}, which comprises 100,000 databases and 1.6 million tables with diverse features. Treating each database as being held by a separate party, we examine cross-party data distributions, identifying four key findings and proposing a data-oriented taxonomy of VFL algorithms. Our review reveals a significant gap between current algorithms and real-world data distributions. We also highlight the challenges in bridging research and practical deployment, offering insights for future research directions.

%% file: sections/overview.tex
In this section, we present the definition of vertical federated learning (VFL) in Section~\ref{subsec:vfl-def} and provide a comprehensive overview of the VFL pipeline. Distinct from previous surveys \cite{liu2024vertical}, which define VFL through the partitioning of a global dataset, we employ a more generalized definition that accounts for the distribution of real-world data.

\subsection{Definition of VFL}\label{subsec:vfl-def}
Consider a collaboration involving \( C \) parties, each possessing a unique and heterogeneous dataset. The dataset of party \( c \) is defined as \( \mathbf{X}^c = \{\mathbf{K}^c, \mathbf{D}^c\} \), where \(\mathbf{K}^c \) denotes the \textit{keys} and \(\mathbf{D}^c \) denotes the \textit{data} associated with party \( c \). Here, the keys \( \mathbf{K}^c \in \mathbb{R}^{n_c \times k} \) are \( k \)-dimensional features shared across all parties, while the data \( \mathbf{D}^c \in \mathbb{R}^{n_c \times d_c} \) represents features specific to each party \( c \).

A VFL task involves multiple parties collaboratively training a machine learning model on the combined datasets \(\{\mathbf{X}^1, \mathbf{X}^2, \ldots, \mathbf{X}^C\}\) while ensuring the privacy of both keys and data. This survey focuses on the widely studied supervised learning scenario where one party possesses the labels, referred to as the \textit{primary party}. The other collaborating parties are termed \textit{secondary parties}. Without loss of generality, we designate $\mathbf{X}^1$ as the primary party. Formally, the VFL task optimizes the following objective function:
\begin{equation}\label{eq:vfl-obj}
\min_{\theta} \frac{1}{n_1} \sum_{i=1}^{n_1} \mathcal{L}(f(\theta; \mathbf{x}_i^1, \mathbf{X}^2, \ldots, \mathbf{X}^C), y_i),
\end{equation}
where $n_1$ is the number of records in the primary party, $\theta$ denotes the model parameters, $\mathcal{L}$ is the loss function, $f$ represents the model, $\mathbf{x}_i^1$ is the $i$-th record from the primary party, and $y_i$ is the label associated with $\mathbf{x}_i^1$.

\subsection{Pipeline of VFL}
The VFL pipeline consists of two main components: \textit{privacy-preserving record linkage} and \textit{VFL training}. The privacy-preserving record linkage component aligns the keys across different parties while safeguarding their privacy. Leveraging this alignment information, the VFL training component collaboratively trains a model on the combined datasets from all parties in a manner that preserves data privacy.

\paragraph{Privacy-Preserving Record Linkage (PPRL).} Privacy-Preserving Record Linkage (PPRL) encompasses a set of techniques designed to align keys across different parties while ensuring the privacy of these keys, such as private set intersection (PSI) \cite{morales2023private}. Formally, given keys $\{\mathbf{K}^c\}_{c=1}^C$ from $C$ parties, PPRL outputs a row selection function $\phi^c$ for each party $c$ such that, for all $c$, each row of $\phi^c(\mathbf{X}^c) \in \mathbb{R}^{n_1 \times m}$ represents the same data instance. In most studies, the row selection function $\phi^1$ for the primary party is typically a constant function, while $\phi^c$ for secondary parties aligns $\mathbf{X}^c$ with the primary party. In the ideal scenario of precise alignment, $\phi^c$ corresponds to multiplying a row permutation matrix. However, in other cases, $\phi^c$ may output a subset of rows or include duplicate rows. Further details will be discussed in Section~\ref{sec:vfl-keys}.

\paragraph{VFL Training.} VFL training refers to the process of collaboratively training a model on aligned datasets without sharing raw data. Formally, building on top of PPRL, VFL training optimizes the following objective function:
\begin{equation}\label{eq:vfl-training-obj}
\min_{\theta} \frac{1}{n_1} \sum_{i=1}^{n_1} \mathcal{L}(f(\theta; \mathbf{x}_i^1, \phi^2(\mathbf{X}^2), \ldots, \phi^C(\mathbf{X}^C)), y_i),
\end{equation}
where $\phi^c$ is the row selection function output by the PPRL component for party $c$. VFL training can be accomplished through various methods. A prevalent approach is split learning, where parties collaborate by exchanging gradients and representations \cite{wang2024unified,nock2021impact}. Alternatively, some methods enable each party to maintain the full set of model parameters and collaborate through boosting strategies \cite{diao2022gal,xian2020assisted}.

%% file: sections/scenario.tex
\begin{figure*}[t!]
    \centering
    \begin{subfigure}[t]{0.24\textwidth}
        \centering
        \includegraphics[width=\textwidth]{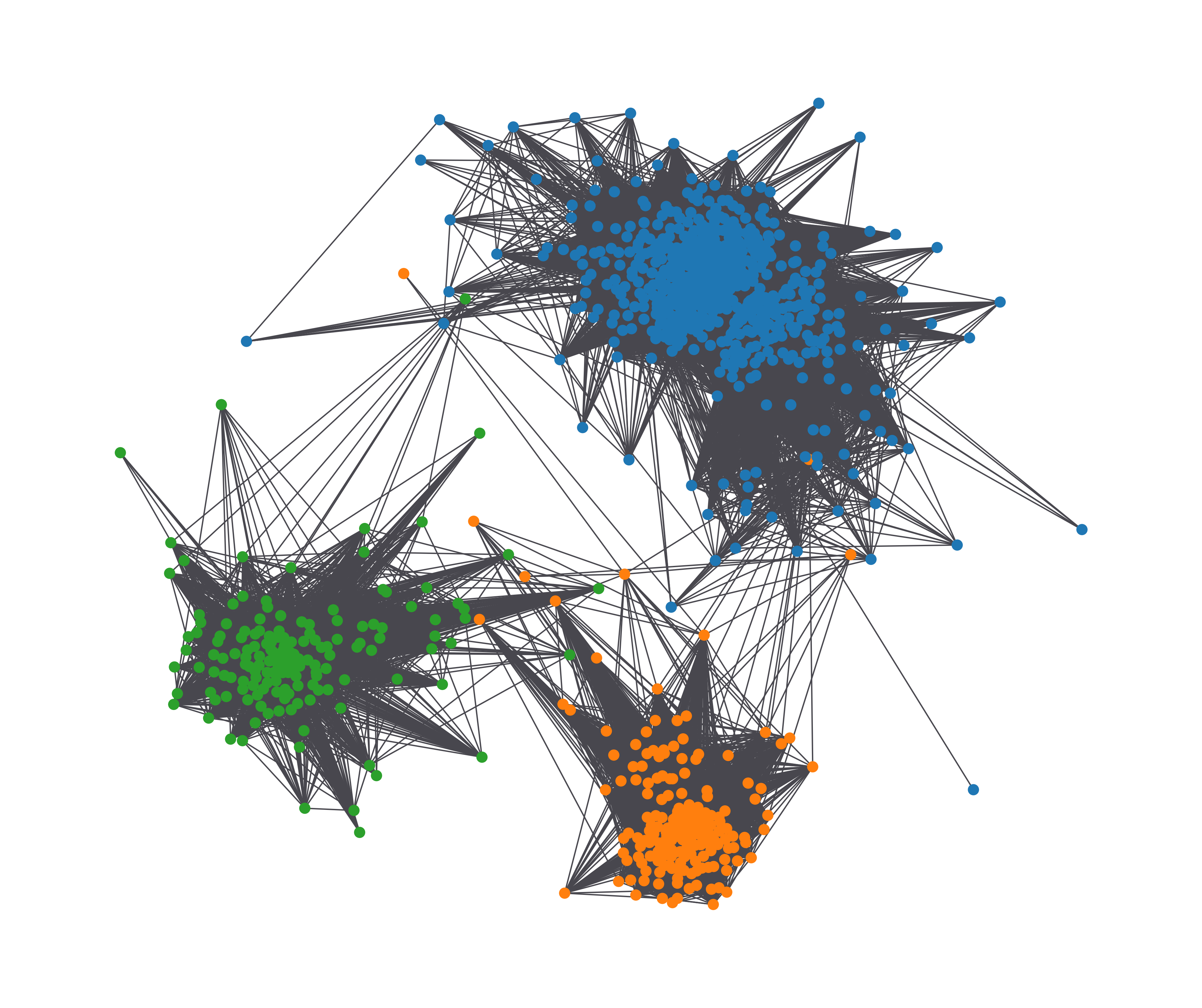}
        \caption{Visualization of 1,000 parties}
        \label{fig:overlap-graph}
    \end{subfigure}
    \hfill
    \begin{subfigure}[t]{0.24\textwidth}
        \centering
        \includegraphics[width=\textwidth]{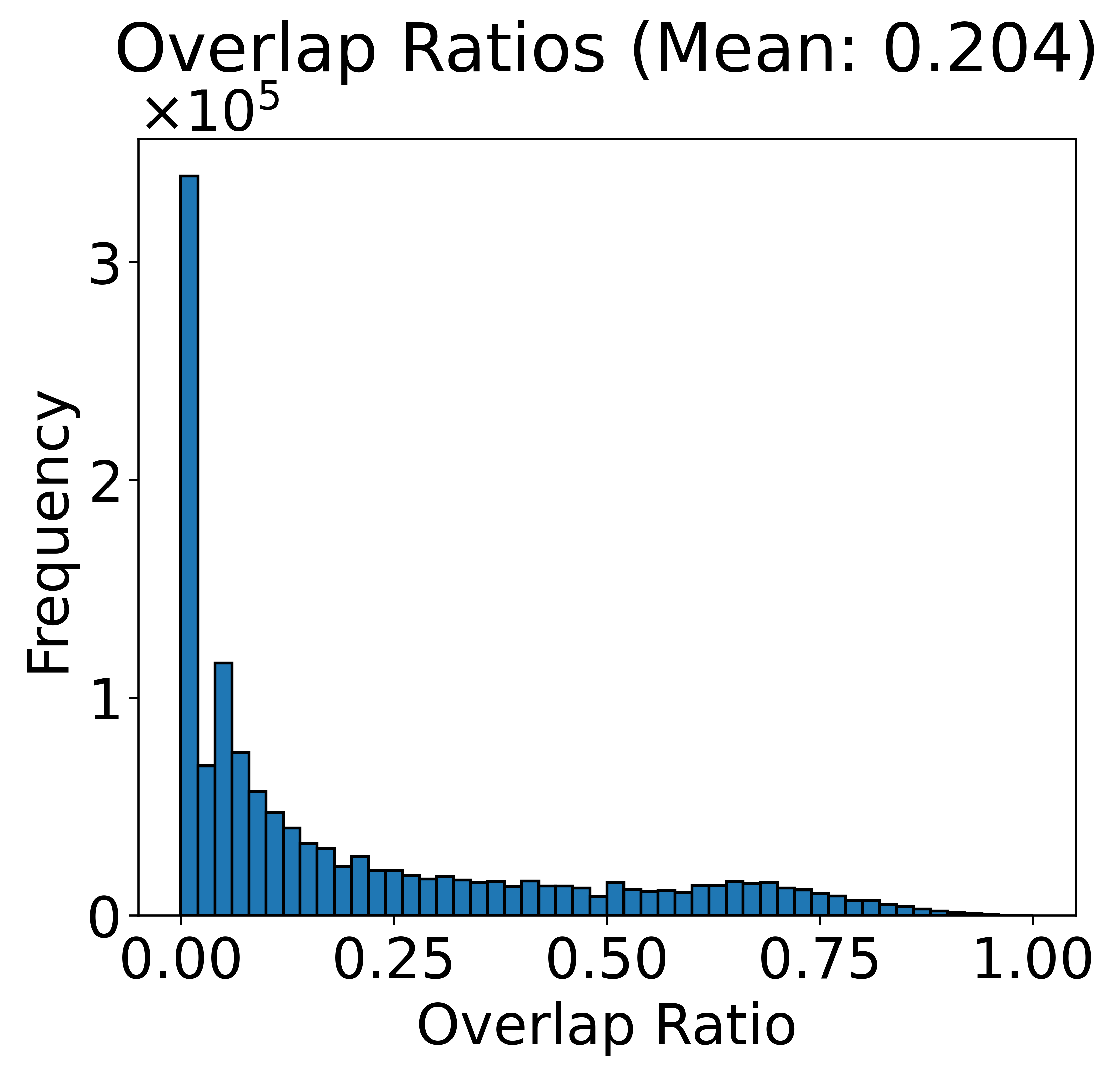}
        \caption{Feature overlap ratios}
        \label{fig:feature-overlap}
    \end{subfigure}
    \hfill
    \begin{subfigure}[t]{0.24\textwidth}
        \centering
        \includegraphics[width=\textwidth]{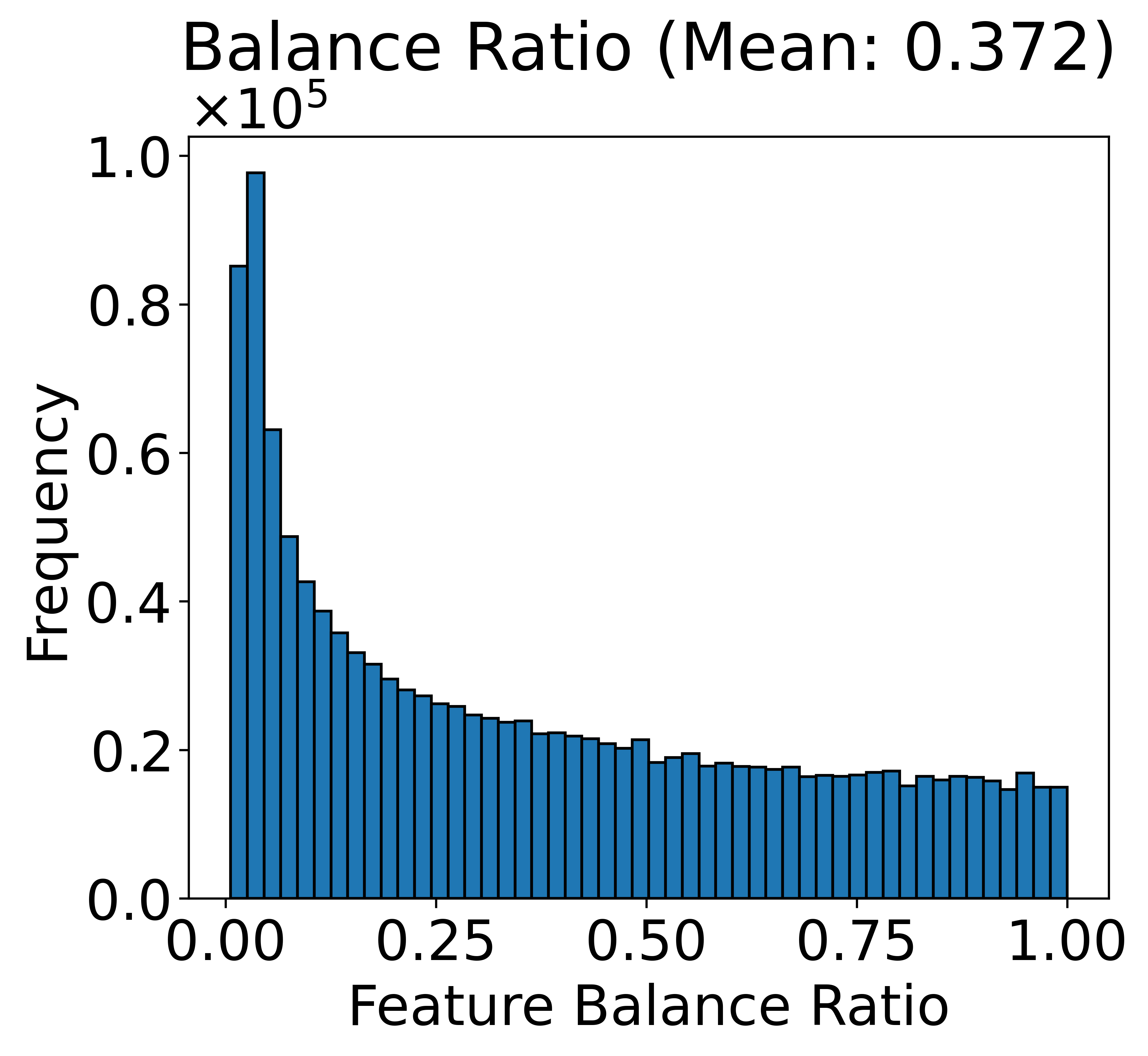}
        \caption{Feature balance ratios}
        \label{fig:feature-skew}
    \end{subfigure}
    \hfill
    \begin{subfigure}[t]{0.24\textwidth}
        \centering
        \includegraphics[width=\textwidth]{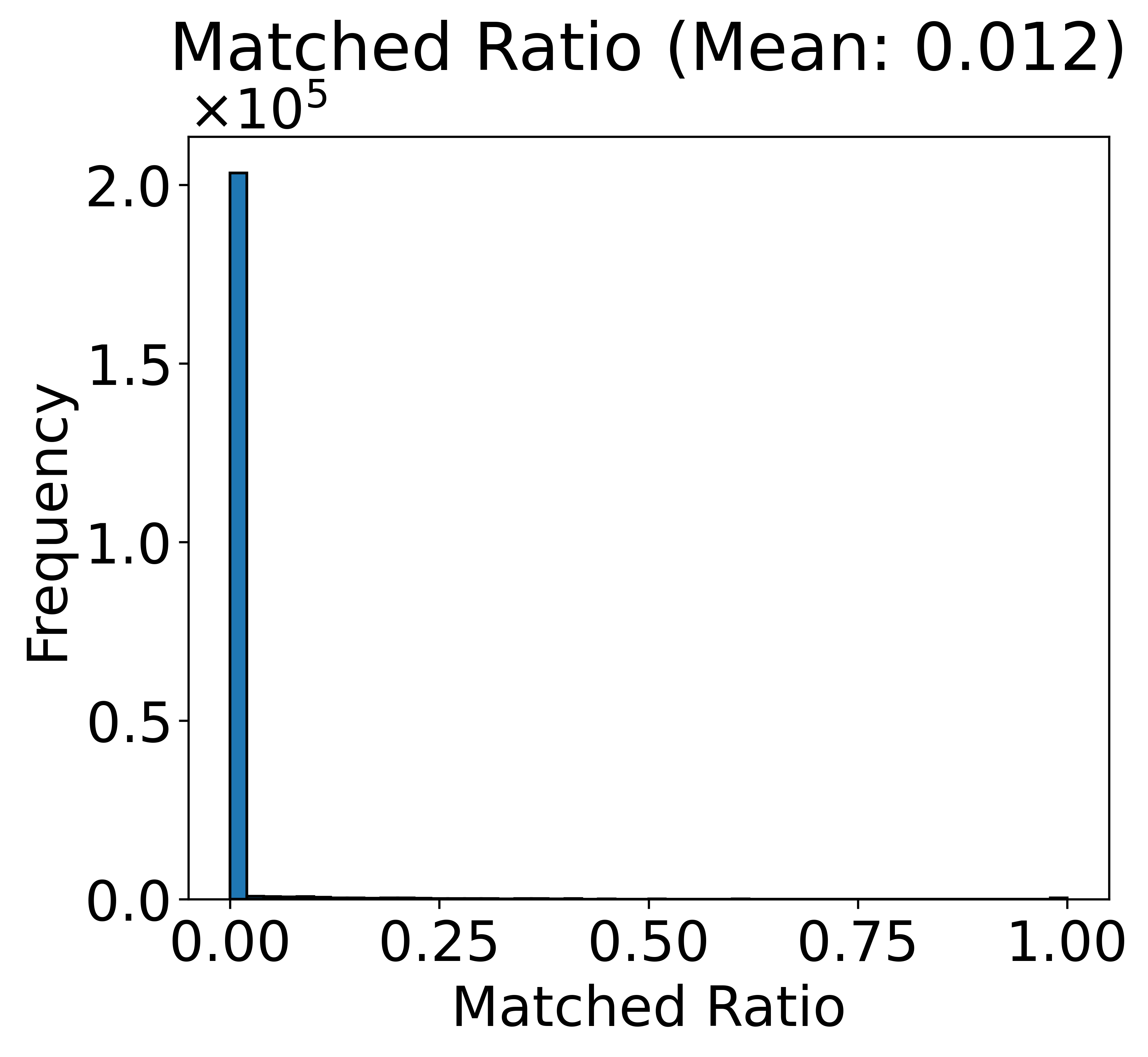}
        \caption{Record matched ratios}
        \label{fig:matched-ratio}
    \end{subfigure}    
    \caption{Analysis and visualization of feature distributions and overlaps across real-world databases}
    \label{fig:feature-analysis}
\end{figure*}

This section examines real-world VFL data distributions using the WikiDBs corpus \cite{vogel2024wikidbs}. We begin by introducing the basic settings in Section~\ref{subsec:settings}, define the essential data properties in Section~\ref{subsec:definitions}, and present our results and findings in Section~\ref{subsec:findings}. 

\subsection{Settings}\label{subsec:settings}
\paragraph{Dataset.}
WikiDBs~\cite{vogel2024wikidbs} is currently the largest publicly available corpus of relational databases, extracted from real-world Wikidata. The details of WikiDBs are shown in Table~\ref{tab:wikidbs}. It covers a wide range of domains, such as clinical, finance, sports, etc. Many of these databases (e.g. \texttt{ucl\_clinical\_research\_trials} and \texttt{apnea\_clinical\_research\_db}) are correlated and can be considered potential scenarios for VFL. Each database can be considered a VFL \textit{party}, with pairs of parties sharing correlated features representing potential VFL pairs. 

\begin{table}[h]
    \centering
    \small
    \caption{Statistics of WikiDBs}
    \label{tab:wikidbs}
    \begin{tabular}{cccc}
        \toprule
        \multicolumn{2}{c}{\textbf{Database}} & \multicolumn{2}{c}{\textbf{Table}} \\
        \cmidrule(lr){1-2} \cmidrule(lr){3-4}
        \#Databases & Total \#Tables & Mean \#Rows & Mean \#Cols \\
        \midrule
        100K & 1.6M & 118 & 52.7 \\
        \bottomrule
    \end{tabular}
\end{table}

\paragraph{Configuration.} In our analysis, we focus on two-party VFL, as it not only reflects the characteristics of multi-party VFL but also represents the most common VFL scenario in practice~\cite{liu2024vertical}. Due to the \(O(n^2)\) complexity of evaluating all pairs, we randomly sampled 1,000 databases, generating 1,000,000 pairs from their Cartesian square. Experiments are repeated across 10 random seeds to reduce variance, with mean and variance reported. The results show consistent subset characteristics, demonstrating the robustness of our analysis.

\subsection{Definitions}\label{subsec:definitions}

This subsection delineates the key properties of VFL data distribution. We begin by defining which pairs of databases are considered potential VFL pairs. At the feature level, we assess their overlap and balance by introducing the \textit{feature overlap ratio} and the \textit{feature balance ratio}. In terms of instances, we quantify the proportion of records that can be matched between two parties using shared features, defining this as the \textit{record matched ratio}. The detailed definitions are as follows.

\paragraph{Potential VFL Pairs.} Potential VFL pairs are defined based on database graph connectivity. In this graph, nodes represent databases, and edges connect nodes with tables sharing at least one column. Two databases are considered potential VFL pairs if they belong to the same connected component, indicating they are related through join operations along the connected path.

\paragraph{Record Matched Ratio.} To evaluate how precisely two VFL parties can be aligned based on shared features, we define the \textit{record matched ratio}. For a given table pair, this ratio measures the fraction of records in each table that identically appear in the other. It ranges from $[0, 1]$ and reflects the proportion of records that can be precisely aligned.

\paragraph{Feature Balance Ratio.} To assess the balance in the number of features between two VFL parties, we define the \textit{feature balance ratio} as the small number of columns between the two databases to the large number of columns. The feature balance ratio ranges from $[0, 1]$, where a value of 1 indicates two parties have the same number of features.

\begin{table}[ht]
    \centering
    \small
    \caption{Properties of the database graph across 10 subsets}
    \label{tab:vfl-properties}
    \setlength{\tabcolsep}{3pt}
    \begin{tabular}{ccccc}
        \toprule
        \textbf{Metric} & \textbf{Mean} & \textbf{Std} & \textbf{Min} & \textbf{Max} \\
        \midrule
        \#Connected Components & 3.1 & 0.3 & 3 & 4 \\
        \midrule
       \makecell{Ratio of Non-Neighboring Nodes \\ in Potential VFL Pairs} & 25.4\% & 1.3\% & 23.7\% & 27.7\% \\
        \bottomrule
    \end{tabular}
\end{table}

\begin{table}[ht]
    \centering
    \small
    \caption{The ratio of different VFL data distributions}
    \label{tab:record-matching}
    \setlength{\tabcolsep}{4pt}
    \begin{tabular}{cccc}
        \toprule
        \textbf{VFL Type} & \textbf{Features} & \textbf{Records} & \textbf{Ratio} \\
        \midrule
        Latent VFL & Zero overlap & Zero match & 25.4\% \\
        \midrule
        Fuzzy VFL & Non-zero overlap & Zero match & 70.9\% \\
        Semi-precise VFL & Non-zero overlap & Partial match & 3.5\% \\
        Precise VFL & Non-zero overlap & Full match & 0.2\% \\
        \bottomrule
    \end{tabular}
\end{table}

\subsection{Results and Findings}\label{subsec:findings}

In this subsection, we present our results in Table~\ref{tab:vfl-properties} and Table~\ref{tab:record-matching}, and Figure~\ref{fig:feature-analysis}, followed by four key findings from our analysis of real-world databases.

\begin{tcolorbox}[colback=white, coltitle=black, boxsep=0mm, left=1mm, right=1mm, top=1.5mm, bottom=1mm]
    \textbf{Finding 1}: VFL has a large number of potential real-world applications.
\end{tcolorbox}
To analyze the connectivity between databases, we constructed a graph where each node represents a database, each edge represents shared features between databases, and each color corresponds to a connected component, as visualized in Figure~\ref{fig:overlap-graph}. Table~\ref{tab:vfl-properties} further summarizes the key properties of this graph, showing that the graph of 1,000 parties contains only 3 to 4 connected components. The high connectivity indicated by both the table and figure suggests that VFL has a wide range of potential real-world applications.

\begin{tcolorbox}[colback=white, coltitle=black, boxsep=0mm, left=1mm, right=1mm, top=1.5mm, bottom=1mm]
\textbf{Finding 2}: A substantial portion (25.4\%) of potential VFL pairs have no overlapping features.
\end{tcolorbox}
To analyze feature overlap ratios between potential VFL pairs, we present the distribution of these ratios in Figure~\ref{fig:feature-overlap} and detail the proportion of pairs with zero feature overlap in Table~\ref{tab:vfl-properties}.  Table~\ref{tab:vfl-properties} reveals that among the potential VFL pairs, approximately one-quarter have zero feature overlap. This indicates that many real-world VFL scenarios require methods that can handle latent relationships without relying on shared features. We define a VFL scenario where two parties have no overlapping features but are still correlated as \textit{latent VFL}.

\begin{tcolorbox}[colback=white, coltitle=black, boxsep=0mm, left=1mm, right=1mm, top=1.5mm, bottom=1mm]
\textbf{Finding 3}: Only a small fraction of potential VFL pairs (0.2\%) can be precisely matched.
\end{tcolorbox}
To evaluate the feasibility of record matching in potential VFL pairs, we analyzed the record matched ratios across database pairs. Figure~\ref{fig:matched-ratio} illustrates the distribution of these ratios, and Table~\ref{tab:record-matching} details the ratios for various VFL data distributions. The results reveal that 70.9\% of potential VFL pairs have zero precise record matches, with only 0.2\% achieving full alignment. This indicates that existing VFL algorithms, which assume fully and precisely matched data records (\textit{precise VFL}), may be inadequate for most real-world applications. In practice, there is a need for algorithms capable of handling partial record matching (\textit{semi-precise VFL}) or even scenarios with no record matching (\textit{fuzzy VFL}). This finding aligns with existing VFL studies~\cite{wu2022coupled,nock2021impact,he2024hybrid}, which highlight the rarity of precise matching.

\begin{tcolorbox}[colback=white, coltitle=black, boxsep=0mm, left=1mm, right=1mm, top=1.5mm, bottom=1mm]
\textbf{Finding 4}: Feature distribution across parties exhibits significant imbalance.
\end{tcolorbox}
To evaluate the balance of features between potential VFL pairs, we present the distribution of feature balance ratio in Figure~\ref{fig:feature-skew}. We observe that most database pairs have highly imbalanced feature counts. This imbalance poses challenges for existing VFL algorithms, which typically assume relatively balanced feature distributions across parties for optimal performance. This finding aligns with the observation in~\cite{wu2023vertibench} that real VFL datasets are highly imbalanced w.r.t. feature importance.Imbalanced VFL is defined as a feature balance ratio below 0.5 (66.49\%), while balanced VFL has a ratio above 0.5 (33.51\%).

%% file: sections/system.tex
\begin{figure*}
    \centering
    \includegraphics[width=\textwidth]{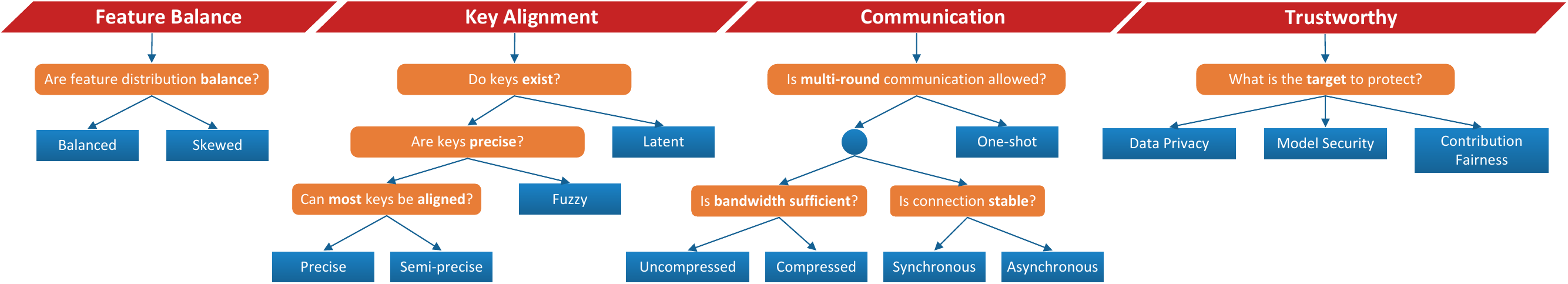}
    \caption{Taxonomy of VFL algorithms based on feature balance, key alignment, communication, and trustworthy.}
    \label{fig:app-guide}
\end{figure*}

Based on the findings in Section~\ref{sec:scenario}, we propose a novel taxonomy of VFL algorithms across four dimensions: keys, features, communication, and trustworthiness, as illustrated in Figure~\ref{fig:app-guide}. Table~\ref{tab:vfl_algorithms} further summarizes existing algorithms according to this taxonomy. The following section provides a detailed explanation of the taxonomy and an overview of the existing algorithms.

\begin{table*}[ht]
    \centering
    \caption{Representative VFL algorithms are categorized by feature balance, key alignment, communication, and trustworthiness. \cmark\;and \xmark\; indicate whether the algorithms include techniques supporting the corresponding VFL category or evaluate it in their experiments. "Contr." denotes contribution fairness. The prevalance of some VFL categories and the ratio of supported algorithms are also presented.}
    \renewcommand{\arraystretch}{1.4}
    \setlength{\tabcolsep}{4pt}
    \resizebox{\textwidth}{!}{
    \begin{tabular}{@{}c|cc|cccc|ccc|ccc@{}}
    \toprule
    \multirow{3}{*}{\textbf{Algorithm}} & \multicolumn{2}{c|}{\textbf{Feature Balance}} & \multicolumn{4}{c|}{\textbf{Key Alignment}} & \multicolumn{3}{c|}{\textbf{Communication}} & \multicolumn{3}{c}{\textbf{Trustworthy}} \\
    \cmidrule(lr){2-3} \cmidrule(lr){4-7} \cmidrule(lr){8-10} \cmidrule(lr){11-13}
    & \makecell{Balanced\\(33.51\%)} & \makecell{Skewed\\(66.49\%)} & \makecell{Precise\\(0.2\%)} & \makecell{Semi-precise\\(3.5\%)} & \makecell{Fuzzy\\(70.9\%)} & \makecell{Latent\\(25.4\%)} & Compressed & Async. & One-shot & Data & Model & Contr. \\
    \midrule
    \cite{hardy2017private} & \cmark & \xmark & \cmark & \xmark & \xmark & \xmark & \xmark & \xmark & \xmark & \cmark & \xmark & \xmark \\
    \cite{chen2020vafl} & \cmark & \xmark & \cmark & \xmark & \xmark & \xmark & \xmark & \cmark & \xmark & \cmark & \xmark & \xmark \\
    \cite{zhang2021asysqn} & \cmark & \cmark & \cmark & \xmark & \xmark & \xmark & \xmark & \cmark & \xmark & \cmark & \xmark & \xmark \\
    \cite{castiglia2022compressed} & \cmark & \xmark & \cmark & \xmark & \xmark & \xmark & \cmark & \xmark & \xmark & \cmark & \xmark & \xmark \\
    \cite{wu2022coupled} & \cmark & \cmark & \cmark & \cmark & \cmark & \xmark & \cmark & \xmark & \xmark & \cmark & \xmark & \xmark \\
    \cite{wu2022practical} & \cmark & \cmark & \cmark & \xmark & \xmark & \xmark & \xmark & \xmark & \cmark & \cmark & \xmark & \xmark \\
    \cite{kang2022fedcvt} & \cmark & \cmark & \cmark & \cmark & \xmark & \xmark & \xmark & \xmark & \xmark & \cmark & \xmark & \xmark \\
    \cite{sun2023communication} & \cmark & \cmark & \cmark & \cmark & \xmark & \xmark & \xmark & \xmark & \cmark & \cmark & \xmark & \xmark \\
    \cite{attack:backdoor:baiyijie2023_villain} & \cmark & \xmark & \cmark & \cmark & \xmark & \xmark & \xmark & \xmark & \xmark & \cmark & \cmark & \xmark \\
    \cite{irureta2024towards} & \cmark & \xmark & \cmark & \cmark & \xmark & \xmark & \cmark & \xmark & \xmark & \cmark & \xmark & \xmark \\
    \cite{sun2024mi} & \cmark & \xmark & \cmark & \xmark & \xmark & \xmark & \cmark & \xmark & \xmark & \cmark & \xmark & \xmark \\
    \cite{wang2024online} & \cmark & \xmark & \cmark & \xmark & \xmark & \xmark & \cmark & \xmark & \xmark & \cmark & \xmark & \xmark \\
    \cite{wang2024computation} & \cmark & \xmark & \cmark & \xmark & \xmark & \xmark & \cmark & \xmark & \xmark & \cmark & \xmark & \xmark \\
    \cite{valdeira2024communication} & \cmark & \xmark & \cmark & \xmark & \xmark & \xmark & \cmark & \xmark & \xmark & \cmark & \xmark & \xmark \\
    \cite{wu2024federated} & \cmark & \cmark & \cmark & \cmark & \cmark & \xmark & \xmark & \xmark & \xmark & \cmark & \xmark & \xmark \\
    \cite{zhang2024asynchronous} & \cmark & \cmark & \cmark & \xmark & \xmark & \xmark & \xmark & \cmark & \xmark & \cmark & \xmark & \xmark \\
    \cite{10.1109/TIFS.2024.3361813} & \cmark & \cmark & \cmark & \xmark & \xmark & \xmark & \xmark & \xmark & \xmark & \cmark & \xmark & \cmark \\
    \cite{attack:dra:zhao_loki_2024} & \cmark & \xmark & \cmark & \xmark & \xmark & \xmark & \xmark & \xmark & \xmark & \cmark & \cmark & \xmark \\
    \cite{attack:backdoor:badvfl} & \cmark & \xmark & \cmark & \xmark & \xmark & \xmark & \xmark & \xmark & \xmark & \cmark & \cmark & \xmark \\
    \cite{attack:backdoor:hijack} & \cmark & \cmark & \cmark & \xmark & \xmark & \xmark & \xmark & \xmark & \xmark & \cmark & \cmark & \cmark \\
    \cite{attack:contribution:ace_2024} & \cmark & \xmark & \cmark & \xmark & \xmark & \xmark & \xmark & \xmark & \xmark & \cmark & \cmark & \xmark \\
    \cite{wang2024unified} & \cmark & \xmark & \cmark & \xmark & \xmark & \xmark & \xmark & \cmark & \xmark & \cmark & \xmark & \xmark \\
    \cite{attack:dra:fu_privacy_2025} & \cmark & \xmark & \cmark & \xmark & \xmark & \xmark & \xmark & \xmark & \xmark & \cmark & \cmark & \xmark \\
    \midrule
    \makecell{Ratio of \;\cmark} & 100\% & 39\% & 100\% & 26\% & 9\% & 0\% & 30\% & 17\% & 9\% & 100\% & 26\% & 9\% \\
    \bottomrule
    \end{tabular}
    }
    \label{tab:vfl_algorithms}
\end{table*}

\subsection{Key Alignment}\label{sec:vfl-keys}

The quality of key alignment is a critical factor in VFL preprocessing. Based on our observations in Section~\ref{sec:scenario}, we categorize VFL frameworks into four types based on their assumptions about key alignment: Precise, Semi-Precise, Fuzzy, and Latent VFL.

\subsubsection{Precise VFL (0.2\%)}

Precise VFL refers to scenarios where all keys are perfectly matched, typically through unique identifiers like user IDs. While uncommon in practice according to Table~\ref{tab:record-matching}, most existing VFL frameworks assume this setting. It applies when each data record corresponds to a user whose ID can be aligned across parties, as seen in some VFL applications using phone numbers as unique IDs~\cite{chen2021homomorphic}. However, precise VFL is rare due to two strict constraints: \textit{all} and \textit{precise}. The \textit{all} constraint is particularly restrictive, requiring both parties to share the exact same set of instances. Studies and benchmarks indicate that cross-party data overlap is generally limited~\cite{qiu2024diverfed,yan2024cross,wu2022coupled,wei2023fedads}.

Precise VFL has been extensively studied and discussed in other surveys \cite{liu2024vertical}. These studies generally follow the PPRL and VFL learning pipeline. Due to the precise key alignment, most research treats PPRL as an orthogonal problem, often addressed by Private Set Intersection (PSI) \cite{morales2023private}, while focusing on the learning aspect. In precise VFL, significant progress has been achieved in various aspects, including accuracy \cite{wang2025pravfed}, efficiency \cite{li2023efficient}, communication \cite{sun2023communication}, and privacy \cite{jin2021cafe}.

\subsubsection{Semi-Precise VFL (3.5\%)}
Semi-precise (a.k.a. semi-supervised) VFL addresses scenarios where only a subset of keys are precisely matched. This scenario encompasses a broader range of real-world applications compared to precise VFL. For instance, in VFL across hospitals, each patient may not undergo all examinations and have records in all hospitals, a situation commonly referred to as ``missing-modal'' in multimodal learning \cite{zong2024self}.

Semi-precise VFL generally follows the PPRL and VFL learning pipeline but incorporates additional methods for handling missing features or label estimation. The specific estimation techniques employed significantly influence the performance of semi-precise VFL. For example, FedCVT \cite{kang2022fedcvt} estimates representations of missing features using scaled dot-product attention (SDPA) and generates pseudo-labels from existing labels. Similarly, \cite{sun2023communication} utilizes analogous techniques but reduces the communication rounds to one or a few. Furthermore, FedAds \cite{wei2023fedads} provides a real-world benchmark for semi-precise VFL. The technical details of semi-precise VFL have been comprehensively summarized in a recent survey \cite{song2024systematic}.

\subsubsection{Fuzzy VFL (70.9\%)}
Fuzzy VFL represents a more challenging scenario where all keys are not precisely matched but the key similarity indicates the probability of match. This situation is common in real-world applications where data records correspond to objects other than identifiable users, such as houses, routes, or products \cite{antoni2019past,wu2022coupled}. In such cases, keys might include GPS coordinates or product descriptions, which are related but not precise enough for direct alignment.

A straightforward approach to fuzzy VFL is to match the top similar pairs of data records and proceed with traditional VFL training \cite{hardy2017private,nock2021impact}. However, these methods often result in suboptimal performance due to the loss of information from less similar pairs. To mitigate this issue, FedSim \cite{wu2022coupled} considers key similarity as a special feature to guide the training of k-nearest-neighbor records, enabling two-party fuzzy VFL. FeT \cite{wu2024federated} further extends fuzzy VFL to multi-party scenarios by encoding keys into representations and employing transformers to identify relationships between these representations. Despite the progress, existing studies on fuzzy VFL still focus on low-dimensional numerical keys, and the performance of high-dimensional keys, such as text embeddings or image features, remains an open problem.

\subsubsection{Latent VFL (25.4\%)}
Latent VFL addresses scenarios where keys are unavailable due to privacy concerns or the absence of shared features. For instance, Singapore's Personal Data Protection Act (PDPA)\footnote{https://www.pdpc.gov.sg/} prohibits clinics from sharing identifiable medical data for research purposes without consent. Another example is a collaboration between a bitcoin transaction company and a bank to detect money laundering. In such cases, because bitcoin wallets are anonymous, there are no keys available to align data across parties. These challenging scenarios necessitate the development of new methods to align data based on their distributions rather than relying on keys. To the best of our knowledge, no existing VFL algorithm supports latent VFL.

\subsection{Feature Balance}\label{sec:vfl-feature-imbalance}
Feature balance is critical for model performance and stability, as shown by a recent benchmark study~\cite{wu2023vertibench}. Figure~\ref{fig:feature-skew} reveals that real-world datasets often exhibit highly imbalanced feature distributions, posing challenges for existing VFL algorithms. Metrics such as Gini impurity~\cite{li2023efficient} and Dirichlet distribution~\cite{wu2023vertibench} evaluate this imbalance. This section categorizes VFL methods into balanced and imbalanced VFL based on a feature balance ratio threshold of 0.5.

\subsubsection{Balanced VFL (33.51\%)}
A balanced feature distribution is ideal for VFL, and most studies~\cite{irureta2024towards,wu2022practical,valdeira2024communication} simulate this by evenly splitting features among parties. However, this setting may not reflect real-world applications. A recent benchmark~\cite{wu2023vertibench} indicates that some methods~\cite{diao2022gal} experience performance degradation under imbalanced feature distributions.

\subsubsection{Imbalance VFL (66.49\%)}
In real VFL scenarios, datasets are distributed across parties with unique and uneven feature characteristics, leading to feature imbalance. This imbalance can significantly affect the performance of VFL \cite{wu2023vertibench}. To address feature imbalance, several algorithms have been developed to mitigate the effects of skewed feature distributions by re-weighting features or adjusting training algorithms. VertiBench~\cite{wu2023vertibench} shows that both SplitNN~\cite{vepakomma2018split} and FedTree~\cite{li2023fedtree} maintain stable performance under imbalanced conditions. A VFL framework~\cite{feng2022vertical} leverages local non-overlapping samples for feature selection, and a Gini impurity-based method~\cite{li2023efficient} accelerates training by filtering out noisy features. Despite these advancements, feature imbalance remains an underexplored challenge in VFL.

\subsection{Communication}
We categorize VFL studies into five types based on communication requirements: one-shot or multi-round VFL, transmission-compressed or uncompressed VFL, and synchronous or asynchronous VFL, as illustrated in Figure~\ref{fig:app-guide}. Since uncompressed and synchronous VFL are the most common, we focus on the remaining three types that address the specific challenges involved with communication.

\subsubsection{One-Shot VFL}
Traditional VFL requires reliable communication channels between the server and the participating clients, which, to some extent, narrows the scope of VFL applications, thereby raising the demand for reducing the number of communication rounds \cite{khan2022communication}. 
The introduction of one-shot VFL has minimized the average number of communication rounds required at the client side \cite{liu2024vertical}. 

Existing works focus on reducing VFL communication costs through unsupervised or semi-supervised learning to generate effective local representations. For example, combining semi-supervised learning with gradient clustering allows clients to learn local feature extractors using overlapping but unaligned samples and share these with the server in a single communication round~\cite{sun2023communication}. FedOnce uses unsupervised learning to generate client-side representations for server-side training in a single round~\cite{wu2022practical}, while APC-VFL integrates representation learning and knowledge distillation to achieve efficient training through aligned sample features~\cite{irureta2024towards}. 

These methods reduce communication rounds by up to hundreds of times, improving VFL's feasibility for clients with limited communication capacity. However, one-shot VFL may impact model accuracy and generalization, particularly under imbalanced data or low sample overlap. Multi-round training allows for gradual optimization, which one-shot approaches may struggle to achieve. Enhancing accuracy through relaxed constraints and few-shot learning algorithms offers a potential solution~\cite{sun2023communication}.

\subsubsection{Transmission-Compressed VFL}
When clients have limited communication budgets, it is common to apply transmission compression between the server and clients, especially for the collaborative training of deep neural networks, which are typically over-parameterized in both their model and feature parameters \cite{xiao2023smoothquant}.

For example, existing works alleviate the burden of data transmission using dimensionality-reduction algorithms \cite{sun2024mi}, quantization \cite{wang2024online}, both quantization and top-k sparsification \cite{castiglia2022compressed}, and pruning applied to both feature model and feature embeddings \cite{wang2024computation}.
Given that existing compression approaches are generally lossy, EFVFL has been proposed to leverage feedback on compression errors, achieving a high convergence rate \cite{valdeira2024communication}.

Although these works demonstrate promising performance in communication efficiency, there is still room for further optimization of data transmission. 
Promising directions include exploring zero-order optimization for VFL and gradient encoding using random seeds to achieve significant reductions in communication overhead \cite{qin24full}.

\subsubsection{Asynchronous VFL}
Traditional synchronous VFL requires all clients to update models simultaneously, creating inefficiencies in practical scenarios where clients have varying computational and network capacities. Asynchronous VFL improves flexibility and efficiency by allowing clients to update local models independently, enhancing the practicality of VFL. 

VAFL~\cite{chen2020vafl} introduces an asynchronous training framework that allows clients to perform stochastic gradient updates independently. AsySQN~\cite{zhang2021asysqn} uses approximate Hessians and a tree-structured collaboration paradigm to reduce communication rounds. VFL-CZOFO~\cite{wang2024unified} applies a hybrid zero-order and first-order optimization, while AVFKAM~\cite{zhang2024asynchronous} asynchronously updates kernel models using pairwise loss functions and random feature approximations. Despite these improvements, asynchronous VFL amplifies client drift~\cite{karimireddy2020scaffold}, as clients may update with outdated global parameters, leading to parameter divergence. Faster clients may dominate training contributions, raising fairness concerns. Addressing these challenges is crucial to ensuring VFL's effectiveness in real deployments.

\subsection{Trustworthy}\label{sec:vfl-attack-model}

VFL applications may involve varying types of sensitive information that require protection. We categorize trustworthiness concerns into three key areas: data privacy, model security, and contribution fairness.

\subsubsection{Data Privacy}
Data privacy is a key concern in VFL, which aims to prevent the exposure of each party's training data. While VFL provides some privacy by transmitting gradients or representations instead of raw data, it does not fully mitigate risks. Effective data protection requires addressing three major vulnerabilities: feature, label, and membership privacy.

\paragraph{Feature.} Existing feature inference attacks~\cite{attack:fia:luo2021} can extract individual features from transmitted embeddings, posing a significant privacy risk. Additionally, data reconstruction attacks can even recreate the original data from these embeddings~\cite{attack:dra:unifv_yang2025uifvdatareconstructionattack}. To mitigate these risks, techniques such as Falcon~\cite{attack:fia:luo2021} and model inconsistency defense~\cite{10.1145/3548606.3560557} should be adopted to secure the aggregation of intermediate embeddings.

\paragraph{Label.} Labels often contain sensitive information that requires protection from unauthorized access. Label inference attacks can deduce labels by analyzing gradient updates during training~\cite{liu2024label}. Advanced techniques like gradient inversion enable attackers to recover labels from batch-averaged local gradients~\cite{attack:lia:liuyang_yangqiang_blackbox}. Both passive and active LIA approaches have been demonstrated, where adversaries can exploit trained bottom models or manipulate gradients to maximize label leakage~\cite{attack:lia:fu2022}. To mitigate these risks, solutions like differential privacy~\cite{abadi2016deep} and encryption-based aggregation~\cite{attack:backdoor:baiyijie2023_villain} are essential to protect labels.

\paragraph{Membership.} While membership inference attacks (MIA) are well-studied in centralized learning to determine whether specific samples were used in training~\cite{attack:mia:survey_bai2024}, their impact on VFL is limited. This is because VFL's mandatory entity alignment process inherently reveals sample membership information between parties~\cite{yu2024survey}, making traditional MIA approaches less effective in the VFL context.

\subsubsection{Model Security}
In real-world VFL applications, adversarial attacks exploit system vulnerabilities, posing risks at both inference and training stages. At inference time, adversaries can craft inputs to induce misclassifications, even with limited access to features or labels~\cite{attack:backdoor:hijack}. These attacks are categorized as either targeted or untargeted. Targeted attacks manipulate specific tasks (e.g., classifying ``striped cars'' as ``birds'' while maintaining overall accuracy)~\cite{attack:backdoor:fool}, whereas untargeted attacks degrade overall model performance by corrupting training data~\cite{attack:backdoor:hijack}. These risks underscore the need for robust security measures to protect both inference and training processes in VFL.

Effective defense methods vary depending on the type of attack. To counter inference-time risks, techniques such as Differential Privacy~\cite{abadi2016deep} help safeguard sensitive data by limiting the information leakage from model outputs. For training-time attacks, including backdoor attacks that embed hidden vulnerabilities into the model~\cite{attack:backdoor:badvfl}, anomaly detection methods such as FreqAnalysis~\cite{safesplit}, originally developed for HFL, show promise in detecting and mitigating malicious updates. By integrating these defense, real-world VFL applications can better ensure both data privacy and model reliability.

\subsubsection{Contribution Fairness}
Protecting intellectual property in VFL requires safeguarding fair contribution evaluation, reward distribution, and preventing attacks like free-riding and contribution fraud. Free-rider attacks involve contributing random or inferred data to obtain a model without meaningful input, making detection difficult~\cite{lin2019freeridersfederatedlearningattacks}. Contribution fraud manipulates evaluation mechanisms to inflate importance and secure unfair rewards, posing a risk in incentive-based systems like Web3 federated learning~\cite{attack:contribution:ofl_w3_linshan}.

Defense mechanisms remain underdeveloped. Free-rider defenses, such as STD-DAGMM~\cite{lin2019freeridersfederatedlearningattacks}, are mainly designed for HFL and are less explored in VFL. Spy~\cite{attack:dra:fu_privacy_2025} highlights new free-rider threats in VFL, but no solution fully prevents these attacks. Fairness-focused methods~\cite{cui2024survey,attack:contribution:vfl_fan2024fair} offer some protection against contribution fraud, though attacks like ACE~\cite{attack:contribution:ace_2024} remain an unaddressed threat. Addressing these gaps is essential to enhance security and fairness in VFL.

%% file: sections/future.tex
Based on our study, we conclude that VFL algorithms are not yet ready for widespread deployment in real-world applications. We identify three critical gaps that must be addressed to enable successful implementation.

\paragraph{Fuzzy and Latent VFL.}  
Many potential VFL applications involve fuzzy or latent data scenarios; however, research in this area remains limited. The primary challenge is to design algorithms capable of handling noise in shared features and uncovering latent relationships within the data. Developing efficient and robust algorithms for these prevalent scenarios would significantly broaden the applicability of VFL.

\paragraph{VFL for Heterogeneous Data.}  
Most existing VFL algorithms are designed for homogeneous data, whereas real-world applications often involve highly heterogeneous data. The key challenge is to create algorithms that can effectively handle data with diverse structures and distributions. Conducting experiments on imbalanced datasets would provide strong evidence of VFL algorithms' effectiveness in practical applications.

\paragraph{Trustworthiness in VFL.}  
Ensuring trustworthiness remains a critical challenge for VFL systems. Our observations indicate that several privacy risks are still unaddressed. Strict privacy and security regulations often impede VFL deployment, as even a single unresolved risk can prevent adoption. Future research must focus on developing comprehensive security mechanisms to enhance the privacy and trustworthiness of VFL systems.

%% file: sections/conclusion.tex
In this study, we investigate the potential applications of VFL and identify four key properties of real VFL data distributions. Based on these findings, we propose a taxonomy and evaluate how existing VFL algorithms address the challenges encountered in practice. Our analysis reveals significant gaps between current VFL algorithms and the demands of real-world applications. We summarize these critical gaps and suggest future research directions to bridge the divide between existing algorithms and practical VFL.